\documentclass{spie}

\usepackage{cite}
\usepackage{hyperref}
\usepackage{times} 
\usepackage{indentfirst} 

\usepackage[colorinlistoftodos,color=pink]{todonotes} 
\usepackage{xkeyval}
\presetkeys{todonotes}{inline}{}

\usepackage{lipsum} 
\usepackage{graphicx}
\usepackage{amsmath}
\usepackage{amssymb}
\usepackage{booktabs}
\usepackage{multirow}
\usepackage {comment}
\usepackage{multirow} 
\usepackage[table]{xcolor}
\usepackage{colortbl}
\newcommand{\wz}[1]{{\textcolor{black}{#1}}}
\definecolor{sampleOne}{RGB}{235,245,255}
\definecolor{sampleThree}{RGB}{235,255,235}
\definecolor{sampleFive}{RGB}{255,245,235}

\title{VT-3DAD: Cross-Category 3D Anomaly Detection via \\ Visual-Text Normal Space Alignment}

\author{Zi Wang\supit{1}, Katsuya Hotta\supit{3}, Yawen Zou\supit{2}, Koichiro Kamide\supit{2}, Yijin Wei\supit{2}, Chao Zhang\supit{2}, Jun Yu$^{\ast}$\supit{1}
  \skiplinehalf
  \normalsize 
  \supit{1}Niigata University, Niigata-city, Japan; \\
  \supit{2}University of Toyama, Toyama-city, Japan; \\
  \supit{3}Iwate University, Morioka, Japan 
  \skiplinehalf
  $^{\ast}$Corresponding author: yujun@ie.niigata-u.ac.jp
}

\begin{document}

\maketitle
\begin{abstract}
Few-shot cross-category 3D anomaly detection aims to determine whether an unknown point cloud belongs to a target normal category using only a few normal references. Existing training-based methods usually require category-wise optimization, while recent training-free methods based on multi-view CLIP visual features mainly rely on visual similarity and may be confused by geometrically similar categories. In this paper, we propose VT-3DAD, a training-free framework for cross-category 3D anomaly detection via Visual-Text Normal Space Alignment. Given few-shot normal references and a test point cloud, VT-3DAD first generates realistic multi-view depth maps and extracts view-wise features using a frozen CLIP visual encoder. The visual branch measures reference-test deviation in the multi-view feature space. In parallel, depth-aware and 3D-aware prompts are encoded by the frozen CLIP text encoder to construct textual normal anchors, which provide semantic normality constraints for the target category. The final anomaly score is obtained by fusing visual deviation from normal references and semantic deviation from the textual normal space. Experiments on the ShapeNetPart dataset demonstrate that VT-3DAD achieves state-of-the-art performance. In particular, VT-3DAD improves the one-shot average AUC-ROC from 92.49\% to 94.80\% compared with the visual-only baseline, while also reducing the average standard deviation from 5.64 to 3.41.

\end{abstract}

\section{Introduction}

Anomaly detection for 3D point clouds is important for industrial inspection, robotic perception, and 3D shape understanding. Most existing 3D anomaly detection methods focus on category-specific defect detection, where both normal and abnormal samples are defined within a fixed object category~\cite{bergmann2021mvtec,liu2023real3d,wang2023multimodal,horwitz2023back}. In this setting, anomalies usually correspond to local geometric defects or surface irregularities. However, many practical scenarios require a more flexible form: given only a few normal point clouds from a target category, the model should determine whether an unknown test sample belongs to the same category or comes from another unseen category. This problem, known as few-shot cross-category 3D anomaly detection, is challenging because the decision boundary must be constructed from very limited normal references, without access to anomalous samples. Existing cross-category methods often rely on reconstruction-based models~\cite{masuda2021toward} or teacher--student frameworks~\cite{qin2023teacher}. Although effective, these methods usually require category-wise training, making them less suitable when the normal category changes frequently.

On the other hand, recent progress in vision-language models has opened a new direction for training-free 3D anomaly detection. CLIP~\cite{radford2021learning} learns transferable visual and textual representations from large-scale image-text pairs, and has been used for 3D recognition by projecting point clouds into multi-view images~\cite{zhu2023pointclip}. Several recent works further explore CLIP-based anomaly detection with multi-view projection or prompt-based adaptation~\cite{cheng2024towards,zuo2024clip3d,zhou2023anomalyclip,ma2025aa}. In particular, DMP-3DAD~\cite{wang2026dmp} shows that realistic depth map projection combined with frozen CLIP visual features can provide a simple and effective training-free solution for cross-category 3D anomaly detection. Nevertheless, visual-only similarity mainly captures projected silhouette cues. When anomalous categories share similar geometries with the target normal category, their multi-view depth maps may produce close visual embeddings, leading to ambiguous anomaly scores. This limitation suggests that visual references alone may be insufficient for robust normality modeling, especially in extreme few-shot settings.

\begin{figure*}[tb]
    \centering
\includegraphics[width=0.8\linewidth]{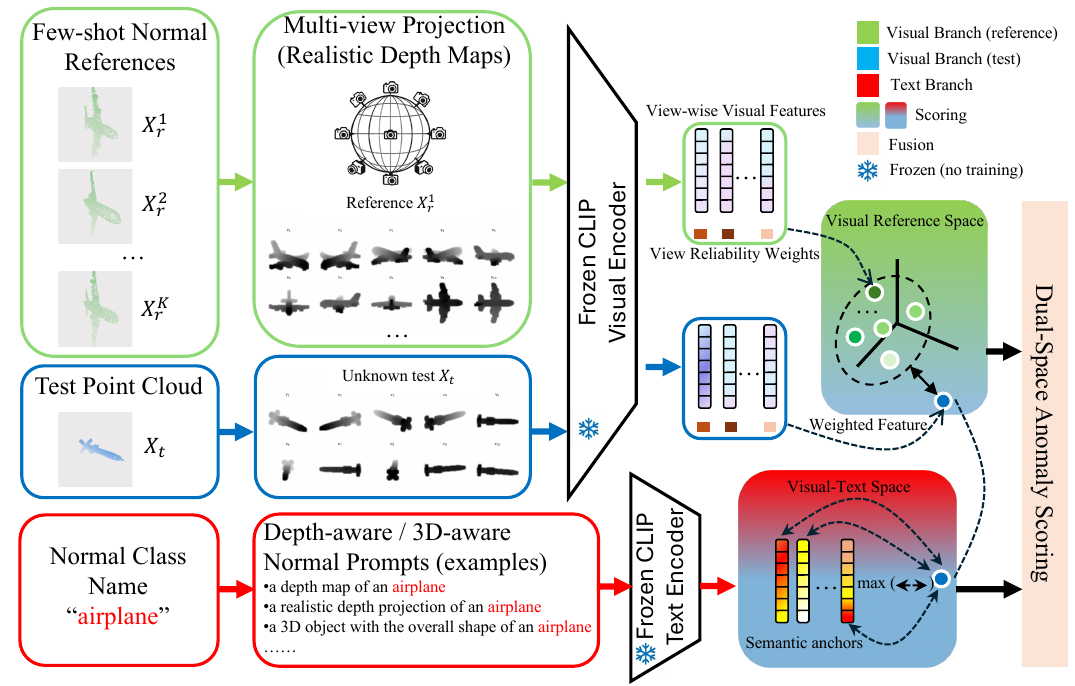}
\caption{Overview of the proposed VT-3DAD framework. Given a few normal reference point clouds and an unknown test point cloud, VT-3DAD first generates realistic multi-view depth maps and extracts view-wise features using a visual encoder. The visual branch measures the weighted view-wise deviation between the test sample and the normal references to obtain a visual deviation score. In parallel, depth-aware and 3D-aware prompts derived from the normal class name are encoded by the text encoder to construct textual semantic anchors. The test visual feature is aligned with these anchors to estimate semantic deviation from the textual normal space. Finally, the visual and semantic deviations are combined through dual-space anomaly scoring to produce the final anomaly score.}
   \label{fig:1}
\end{figure*}

To address this issue, we consider the text branch of CLIP from a normality modeling perspective. Instead of using text prompts as a zero-shot classifier, we use them to construct a semantic normal concept space for the target category. Specifically, depth-aware and 3D-aware textual prompts describe the normal category in a way that is consistent with multi-view depth projections, and their CLIP text embeddings serve as semantic anchors. These anchors provide category-level normal concepts that are complementary to instance-level visual references. Based on this idea, we propose VT-3DAD, a training-free framework via Visual-Text Normal Space Alignment. As illustrated in Fig.~\ref{fig:1}, the framework measures anomaly from two spaces: the visual reference space, which captures deviation from few-shot normal samples, and the textual normal space, which captures semantic inconsistency with the target normal concept. By combining these two signals, VT-3DAD improves both discrimination and stability without introducing training, fine-tuning, or anomalous samples. The main contributions of this work are summarized as follows:
\begin{itemize}
    \item We propose VT-3DAD, a training-free framework for few-shot cross-category 3D anomaly detection that aligns visual references with textual semantic concepts to model normality.
    \item We construct a textual normal concept space with frozen CLIP text features from depth-aware and 3D-aware prompts, using text as semantic anchors.
    \item We introduce a dual-space anomaly score that combines visual deviation from normal references and semantic deviation from the textual normal space, improving robustness to geometrically similar anomalies.
    \item Experiments on ShapeNetPart show that our method achieves state-of-the-art performance under 1-, 3-, and 5-shot settings, improving the 1-shot average AUC-ROC from 92.49\% to 94.80\% compared with DMP-3DAD.
\end{itemize}

\section{Method}

\subsection{Problem Setting}

We study few-shot cross-category 3D anomaly detection for point clouds. Let $\mathcal{C}$ denote the set of object categories and let $c^\ast \in \mathcal{C}$ be the target normal category. Given a point cloud $X \in \mathbb{R}^{N \times 3}$ with $N$ points, samples from $c^\ast$ are regarded as normal, while samples from any other category $c \in \mathcal{C}\setminus\{c^\ast\}$ are treated as anomalies. During reference construction, only a small set of normal point clouds is available:
\begin{equation}
    \mathcal{R}=\{X_r^i\}_{i=1}^{K}, \quad X_r^i \sim p(X|c^\ast),
\end{equation}
where $K$ is the number of few-shot normal references. No anomalous samples are used. At test time, an unknown point cloud $X_t$ is assigned an anomaly score $s(X_t)$, where a larger value indicates a higher likelihood of being anomalous. Our framework is training-free. For each target normal category, VT-3DAD only extracts frozen CLIP features from the few-shot references and the test point clouds, constructs textual normal anchors from the normal class name, and computes anomaly scores without optimizing any learnable parameters.

\subsection{Realistic Multi-view Projection}
Directly applying 2D pretrained models to raw point clouds is difficult because of the domain gap between irregular 3D points and 2D images. Following the realistic projection strategy \cite{zhu2023pointclip}, each point cloud is converted into a fixed number of depth-like images from predefined viewpoints. Specifically, given a point cloud $X$, the projection module produces $V$ rendered views:
\begin{equation}
    \mathcal{I}(X)=\{I_1(X), I_2(X), \ldots, I_V(X)\},
\end{equation}
where $V=10$ in our implementation. Each projected image is resized to the input resolution of the CLIP visual encoder. Since the projection process is deterministic and contains no learnable parameters, it preserves the training-free nature of the proposed framework. Also, following \cite{wang2026dmp}, not all views are equally informative. Some views may contain more foreground pixels and thus provide more reliable geometric evidence. Therefore, we estimate a view reliability weight from the few-shot normal references. For a projected image $I_v(X_r^i)$, a pixel is regarded as foreground if at least one of its channels has an intensity smaller than a threshold $\gamma$. In our implementation, $\gamma=0.2$. Let $M_{i,v}(x,y)$ be the resulting binary foreground mask. The foreground ratio of the $v$-th view for the $i$-th reference sample is computed as
\begin{equation}
    r_{i,v}=\frac{1}{HW}\sum_{x=1}^{H}\sum_{y=1}^{W}M_{i,v}(x,y),
\end{equation}
where $H$ and $W$ denote the height and width of the projected image. The final view reliability weight is obtained by averaging over all normal references:
\begin{equation}
    w_v=\frac{1}{K}\sum_{i=1}^{K}r_{i,v}.
\end{equation}
The weights $\{w_v\}_{v=1}^{V}$ are used in both the visual reference matching branch and the visual-text alignment branch.

\subsection{Visual Reference Space and Visual Deviation}
Each projected view is encoded by the frozen CLIP visual encoder. Let $E_V(\cdot)$ denote the CLIP visual encoder. For the $v$-th view of point cloud $X$, the view-wise visual feature is
\begin{equation}
    f_v(X)=\frac{E_V(I_v(X))}{\|E_V(I_v(X))\|_2}.
\end{equation}
The multi-view representation of $X$ is then written as
\begin{equation}
    F(X)=\{f_1(X),f_2(X),\ldots,f_V(X)\}.
\end{equation}
The visual branch measures how far the test sample deviates from the few-shot normal references in the multi-view visual feature space. For a test sample $X_t$ and a reference sample $X_r^i$, we compute the weighted view-wise distance as
\begin{equation}
    d_v(X_t,X_r^i)
    =
    \sum_{v=1}^{V}
    \left\|
    w_v f_v(X_t)-w_v f_v(X_r^i)
    \right\|_2 .
\end{equation}
The visual deviation score is obtained by summing the distances to all few-shot normal references:
\begin{equation}
    s_v(X_t)=\sum_{i=1}^{K}d_v(X_t,X_r^i).
\end{equation}
A larger $s_v(X_t)$ indicates that the test point cloud is visually farther from the normal reference space and is therefore more likely to be anomalous.

\subsection{Textual Normal Concept Space}

The visual branch relies on few-shot normal references and may be sensitive to the selected references, especially when $K$ is small. To provide a complementary semantic cue, we construct a textual normal concept space from the target normal class name using the frozen CLIP text encoder. For a normal category $c^\ast$, we first build a set of $M$ textual prompts:
\begin{equation}
    \mathcal{P}_{c^\ast}=\{p_m(c^\ast)\}_{m=1}^{M}.
\end{equation}
In the default semantic prompt set, the prompts are depth-aware and 3D-aware, such as ``a depth map of a \textit{class}'', ``a realistic depth projection of a \textit{class}'', and ``a point cloud object belonging to the \textit{class} category''. These prompts are designed to better match the multi-view depth projection inputs used by the visual branch. Let $E_T(\cdot)$ denote the CLIP text encoder. Each prompt is encoded and normalized as
\begin{equation}
    t_m=
    \frac{E_T(p_m(c^\ast))}
    {\|E_T(p_m(c^\ast))\|_2},
    \quad m=1,\ldots,M.
\end{equation}
The resulting text embeddings form the textual normal concept space:
\begin{equation}
    \mathcal{T}_{c^\ast}=\{t_1,t_2,\ldots,t_M\}.
\end{equation}
Different from zero-shot classification, we do not use text prompts to classify among multiple categories. Instead, the text embeddings are used as semantic anchors that describe the normal concept of the target category.

\subsection{Visual-Text Normal Space Alignment}

To compare the test point cloud with the textual normal concept space, the multi-view visual features of the test sample are aggregated into a single global feature. Since text prototypes do not have a view dimension, we use view-weighted mean pooling:
\begin{equation}
    \bar{f}(X_t)
    =
    \frac{
    \sum_{v=1}^{V}\tilde{w}_v f_v(X_t)
    }
    {
    \left\|
    \sum_{v=1}^{V}\tilde{w}_v f_v(X_t)
    \right\|_2
    },
\end{equation}
where
\begin{equation}
    \tilde{w}_v=\frac{w_v}{\sum_{j=1}^{V}w_j}.
\end{equation}
The semantic normality of the test sample is measured by the maximum cosine similarity between the global visual feature and the text prototypes:
\begin{equation}
    s_\tau(X_t)
    =
    \max_{m}
    \bar{f}(X_t)^\top t_m .
\end{equation}
A larger $s_\tau(X_t)$ means that the test sample is semantically more consistent with the textual normal concept space. Therefore, its semantic deviation can be written as $1-s_\tau(X_t)$ after score normalization.

\subsection{Dual-Space Anomaly Scoring}

The visual branch and text branch produce scores with different numerical ranges. Therefore, before fusion, we linearly rescale both the visual deviation score and the textual normality score to $[0,1]$ over the test set:
\begin{equation}
    \hat{s}=\frac{s-\min(s)}{\max(s)-\min(s)}.
\end{equation}
Let $\hat{s}_v(X_t)$ denote the normalized visual deviation score and $\hat{s}_\tau(X_t)$ denote the normalized textual normality score. The final anomaly score is computed as
\begin{equation}
    s(X_t)
    =
    \alpha \hat{s}_v(X_t)
    +
    \beta \left(1-\hat{s}_\tau(X_t)\right),
\end{equation}
where $\alpha$ and $\beta$ control the contributions of the visual and text branches, respectively. In our default setting, we use $\alpha=1.0$ and $\beta=0.5$. A test sample receives a high anomaly score when it is visually far from the few-shot normal references or semantically inconsistent with the textual normal space.

\section{Experiment}
\subsection{Experimental Setup}

\noindent\textbf{Dataset.}
Following previous cross-category 3D anomaly detection studies~\cite{masuda2021toward,qin2023teacher,wang2026dmp}, we evaluate the proposed VT-3DAD on the ShapeNetPart dataset~\cite{yi2017large}. ShapeNetPart contains object-level point clouds from 16 semantic categories, including airplane, chair, table, lamp, laptop, rocket, and other common object classes. Each object is represented as a point cloud sampled from the corresponding 3D shape surface. In our experiments, each category is alternately selected as the target normal category, while samples from all other categories are regarded as cross-category anomalies.

\noindent\textbf{Few-shot protocol.}
We consider the few-shot one-class setting, where only a small number of normal samples from the target category are available as references. Specifically, we evaluate three settings with $K \in \{1,3,5\}$ normal reference samples. No anomalous samples are used during reference construction. For each target category, the anomaly detector assigns an anomaly score to each test sample, and the score is expected to be lower for samples from the target normal category and higher for samples from other categories.

\noindent\textbf{Evaluation metric.}
We use the area under the receiver operating characteristic curve (AUC-ROC) as the evaluation metric. AUC-ROC is computed independently for each target category, and the average over all categories is reported as the final performance. To reduce the influence of random reference selection, all experiments are repeated with 10 random seeds. We report both the mean AUC-ROC and the average category-wise standard deviation over 10 runs.

\noindent\textbf{Implementation details.}
Unless otherwise specified, we use CLIP ViT-B/32 as the visual and textual backbone. All CLIP encoders are frozen, and no model training or fine-tuning is performed. Each point cloud is projected into $V=10$ realistic depth maps from predefined viewpoints. The view reliability threshold is set to $\gamma=0.2$. For the dual-space anomaly score, we set $\alpha=1.0$ and $\beta=0.5$ by default. The semantic prompt set is used as the default textual prompt design.

\begin{table*}[t]
    \centering
    \caption{Category-wise \wz{AUC-ROC} (\%) comparison for few-shot anomaly detection.
The target category (normal class) is used for reference, and all other categories are regarded as anomalies. Results are reported with 1, 3, and 5 reference samples. Mean and standard deviation are computed over 10 random runs. $\uparrow$ ($\downarrow$) indicates that higher (lower) values are better.
Best results are shown in bold.}
    \label{tab:tab2}
    
    \setlength{\tabcolsep}{3pt}
    \renewcommand{\arraystretch}{1.2}
    \tiny
    
    \begin{tabular}{
l
>{\columncolor{sampleOne}}c
>{\columncolor{sampleThree}}c
>{\columncolor{sampleFive}}c
>{\columncolor{sampleOne}}c
>{\columncolor{sampleThree}}c
>{\columncolor{sampleFive}}c
>{\columncolor{sampleOne}}c
>{\columncolor{sampleThree}}c
>{\columncolor{sampleFive}}c
>{\columncolor{sampleOne}}c
>{\columncolor{sampleThree}}c
>{\columncolor{sampleFive}}c
}
        \toprule
        & \multicolumn{3}{c}{Reconstruction-based~\cite{masuda2021toward}}
        & \multicolumn{3}{c}{Knowledge-distillation-based~\cite{qin2023teacher}}
        & \multicolumn{3}{c}{DMP-3DAD\cite{wang2026dmp}}
        & \multicolumn{3}{c}{VT-3DAD (Ours)} \\
        \cmidrule(lr){2-4}
        \cmidrule(lr){5-7}
        \cmidrule(lr){8-10}
        \cmidrule(lr){11-13}
        \multicolumn{1}{c}{Category (\#tests)}
        & 1 sample & 3 samples & 5 samples
        & 1 sample & 3 samples & 5 samples
        & 1 sample & 3 samples & 5 samples
        & 1 sample & 3 samples & 5 samples \\
        \midrule
        Airplane (341)
        & 87.60 $\pm$ 5.99  & 96.07 $\pm$ 1.07  & 97.39 $\pm$ 0.48
        & 97.41 $\pm$ 1.01  & 98.59 $\pm$ 0.18 & 98.29 $\pm$ 0.53
        & 98.73 $\pm$ 2.22 & 99.71 $\pm$ \textbf{0.11} & 99.75 $\pm$ \textbf{0.08}
        & \textbf{99.56}$\pm$\textbf{0.47} & \textbf{99.72}$\pm$\textbf{0.11} & \textbf{99.77}$\pm$0.09 \\
        
        Bag (14)
        & 47.09 $\pm$ 8.73  & 52.41 $\pm$ 5.86  & 58.70 $\pm$ 5.79
        & \textbf{96.88} $\pm$ \textbf{3.73}  & \textbf{98.23} $\pm$ \textbf{2.13}  & \textbf{99.94} $\pm$ \textbf{0.08}
        & 90.26$\pm$6.57 & 94.18$\pm$3.98 & 94.85$\pm$3.51
        & 94.80$\pm$4.08 & 96.48$\pm$2.40 & 97.04$\pm$1.55 \\
        
        Cap (11)
        & 38.71 $\pm$ 5.70  & 45.22 $\pm$ 3.82  & 46.31 $\pm$ 6.38
        & 90.96 $\pm$ 5.54  & 94.96 $\pm$ 3.05 & 94.13 $\pm$ 2.94
        & 97.85$\pm$1.30 & 98.35$\pm$1.35 & 98.18$\pm$2.20
        & \textbf{99.42}$\pm$\textbf{0.88} & \textbf{99.42}$\pm$\textbf{0.88} & \textbf{99.26}$\pm$\textbf{0.91} \\
        
        Car (158)
        & 62.28 $\pm$ 3.27  & 64.12 $\pm$ 2.06  & 65.14 $\pm$ 1.81
        & 99.33 $\pm$ \textbf{0.27}  & 99.36 $\pm$ 0.26 & 99.31 $\pm$ 0.29
        & 99.32$\pm$0.69 & 99.69$\pm$0.14 & 99.75$\pm$0.09
        & \textbf{99.62}$\pm$0.33 & \textbf{99.78}$\pm$\textbf{0.11} & \textbf{99.81}$\pm$\textbf{0.06} \\
        
        Chair (704)
        & 49.20 $\pm$ 4.27  & 53.38 $\pm$ 2.96  & 55.38 $\pm$ 1.29
        & \textbf{95.16} $\pm$ \textbf{2.20}  & \textbf{98.54} $\pm$ \textbf{0.64}  & \textbf{98.72} $\pm$ \textbf{0.18}
        & 84.34$\pm$7.56 & 93.10$\pm$3.52 & 93.33$\pm$2.30
        & 88.03$\pm$3.69 & 92.32$\pm$2.13 & 92.30$\pm$1.89 \\
        
        Earphone (14)
        & 38.78 $\pm$ \textbf{7.61}  & 45.36 $\pm$ 7.26  & 43.64 $\pm$ 3.21
        & 81.97 $\pm$ 23.45 & 91.31 $\pm$ 3.07 & 90.19 $\pm$ 2.22
        & 89.08$\pm$21.45 & \textbf{99.95}$\pm$\textbf{0.16} & \textbf{100.00}$\pm$\textbf{0.00}
        & \textbf{92.76}$\pm$13.39 & \textbf{99.95}$\pm$\textbf{0.16} & 99.95$\pm$0.16 \\
        
        Guitar (159)
        & 71.75 $\pm$ 3.45  & 76.13 $\pm$ 3.74  & 77.59 $\pm$ 2.57
        & \textbf{98.65} $\pm$ \textbf{0.54} & 97.66 $\pm$ 1.13 & 98.39 $\pm$ 0.69
        & 97.53$\pm$1.93 & \textbf{98.99}$\pm$\textbf{0.86} & \textbf{99.22}$\pm$\textbf{0.30}
        & 96.13$\pm$2.32 & 98.10$\pm$1.04 & 98.53$\pm$0.45 \\
        
        Knife (80)
        & 66.46 $\pm$ 4.20  & 70.49 $\pm$ 2.31  & 71.79 $\pm$ \textbf{0.91}
        & \textbf{95.18} $\pm$ \textbf{1.75}  & \textbf{95.33} $\pm$ \textbf{2.22}  & \textbf{96.72} $\pm$ 1.08
        & 88.09$\pm$15.13 & 91.84$\pm$4.18 & 93.74$\pm$2.69
        & 91.53$\pm$5.95 & 92.83$\pm$3.12 & 94.28$\pm$1.79 \\
        
        Lamp (286)
        & 53.09 $\pm$ \textbf{5.02}  & 58.68 $\pm$ \textbf{2.55}  & 62.20 $\pm$ 3.68
        & 56.13 $\pm$ 6.16  & 60.76 $\pm$ 8.34  & 61.22 $\pm$ 5.84
        & 68.48$\pm$11.70 & 82.65$\pm$5.95 & 83.20$\pm$2.90
        & \textbf{80.23}$\pm$7.06 & \textbf{88.30}$\pm$3.23 & \textbf{87.94}$\pm$\textbf{1.85} \\
        
        Laptop (83)
        & 67.36 $\pm$ 4.52  & 69.08 $\pm$ 2.24  & 70.24 $\pm$ 2.58
        & 98.89 $\pm$ 0.16 & 98.78 $\pm$ 0.32 & 98.69 $\pm$ 0.29
        & 98.88$\pm$\textbf{0.05} & 98.89$\pm$\textbf{0.06} & 98.90$\pm$\textbf{0.05}
        & \textbf{99.30}$\pm$0.11 & \textbf{99.30}$\pm$0.09 & \textbf{99.28}$\pm$0.08 \\
        
        Motorbike (51)
        & 82.55 $\pm$ 2.75  & 87.66 $\pm$ 1.28  & 88.09 $\pm$ 2.56
        & 90.63 $\pm$ 26.57 & 99.27 $\pm$ 0.72 & 99.35 $\pm$ 0.60
        & 98.19$\pm$3.11 & 99.62$\pm$0.66 & 99.68$\pm$0.52
        & \textbf{99.59}$\pm$\textbf{0.69} & \textbf{99.97}$\pm$\textbf{0.04} & \textbf{99.97}$\pm$\textbf{0.03} \\
        
        Mug (38)
        & 44.04 $\pm$ 5.67  & 46.94 $\pm$ 2.44  & 48.55 $\pm$ 3.82
        & 99.60 $\pm$ 0.49  & 99.73 $\pm$ 0.37 & 99.35 $\pm$ 0.57
        & 99.68$\pm$0.50 & 99.96$\pm$0.06 & 99.96$\pm$0.06
        & \textbf{99.99}$\pm$\textbf{0.03} & \textbf{100.00}$\pm$\textbf{0.00} & \textbf{100.00}$\pm$\textbf{0.00} \\
        
        Pistol (44)
        & 64.85 $\pm$ 4.70  & 71.77 $\pm$ 4.41  & 78.57 $\pm$ 4.44
        & 99.08 $\pm$ \textbf{0.66} & 98.86 $\pm$ 0.51 & 98.80 $\pm$ 0.48
        & 98.67$\pm$1.43 & 99.41$\pm$1.14 & 99.53$\pm$0.65
        & \textbf{99.60}$\pm$0.81 & \textbf{99.87}$\pm$\textbf{0.26} & \textbf{99.94}$\pm$\textbf{0.10} \\
        
        Rocket (12)
        & 55.25 $\pm$ 2.97  & 55.95 $\pm$ 5.69  & 59.67 $\pm$ 5.05
        & \textbf{96.63} $\pm$ \textbf{1.44} & 95.73 $\pm$ 2.01 & 96.23 $\pm$ 2.93
        & 89.38$\pm$6.94 & 93.68$\pm$4.98 & 94.24$\pm$4.10
        & 95.35$\pm$5.65 & \textbf{97.78}$\pm$\textbf{1.53} & \textbf{98.61}$\pm$\textbf{1.13} \\
        
        Skateboard (31)
        & 47.84 $\pm$ 5.19  & 57.56 $\pm$ 5.08  & 57.29 $\pm$ 6.62
        & \textbf{96.16} $\pm$ \textbf{1.20} & 95.64 $\pm$ \textbf{1.06} & 96.07 $\pm$ \textbf{1.11}
        & 94.59$\pm$3.66 & \textbf{97.17}$\pm$2.42 & \textbf{98.11}$\pm$1.37
        & 90.03$\pm$5.90 & 93.69$\pm$2.73 & 94.72$\pm$1.73 \\
        
        Table (848)
        & 47.39 $\pm$ 14.48 & 61.97 $\pm$ 14.66 & 78.78 $\pm$ 4.56
        & 83.46 $\pm$ 8.17  & 89.77 $\pm$ 2.34  & 90.40 $\pm$ \textbf{1.05}
        & 86.82$\pm$5.98 & 90.03$\pm$1.90 & 90.65$\pm$2.00
        & \textbf{90.86}$\pm$\textbf{3.21} & \textbf{92.19}$\pm$\textbf{1.71} & \textbf{92.58}$\pm$1.73 \\
        
        \midrule
        Avg. \wz{AUC-ROC} $\uparrow$
        & 57.77 & 63.30 & 66.21
        & 92.26 & 94.53 & 94.74
        & 92.49 & 96.08 & 96.44
        & \textbf{94.80} & \textbf{96.86} & \textbf{97.12} \\
        
        Avg. Std. Dev. $\downarrow$
        & 5.53  & 4.21  & 3.48
        & 5.21  & 1.77  & 1.30
        & 5.64  & 1.97  & 1.43
        & \textbf{3.41} & \textbf{1.22} & \textbf{0.85} \\
        \bottomrule
    \end{tabular}
\end{table*}

\subsection{Comparison with State-of-the-Art Methods}

Table~\ref{tab:tab2} compares VT-3DAD with representative few-shot cross-category 3D anomaly detection methods, including a reconstruction-based method~\cite{masuda2021toward}, a knowledge-distillation-based method~\cite{qin2023teacher}, and the visual-only DMP-3DAD baseline~\cite{wang2026dmp}. The reconstruction-based and knowledge-distillation-based methods require category-wise training using the normal samples. Overall, VT-3DAD achieves the best average performance under all few-shot settings. Specifically, our method obtains average AUC-ROC scores of 94.80\%, 96.86\%, and 97.12\% with 1, 3, and 5 normal reference samples, respectively. Compared with DMP-3DAD, VT-3DAD improves the average AUC-ROC from 92.49\% to 94.80\% in the 1-shot setting, from 96.08\% to 96.86\% in the 3-shot setting, and from 96.44\% to 97.12\% in the 5-shot setting. The improvement is most evident in the 1-shot setting, indicating that textual normal anchors provide useful semantic priors when visual normal references are extremely limited. In addition to improving the average AUC-ROC, VT-3DAD also reduces the sensitivity to random reference selection. Compared with DMP-3DAD, the average standard deviation decreases from 5.64 to 3.41 in the 1-shot setting, from 1.97 to 1.22 in the 3-shot setting, and from 1.43 to 0.85 in the 5-shot setting. This demonstrates that the textual normal concept space provides a stable semantic reference and improves the robustness of anomaly scoring.


\subsection{Ablation on Fusion Weights}

We investigate the influence of the fusion weights in the dual-space anomaly score. As shown in Table~\ref{tab:ablation_alpha_beta}, we fix $\alpha=1.0$ and vary $\beta$ from 0.25 to 1.50. When $\beta$ is small, the method mainly relies on visual reference matching, and the contribution of the textual normal space is limited. Increasing $\beta$ is especially helpful in the 1-shot setting, where the single normal reference is randomly selected and may not reliably represent the target category. In this case, textual normal anchors provide a more stable semantic prior, leading to improved performance and reduced standard deviation. Although $\beta=1.0$ slightly improves the 1-shot mean AUC-ROC, it degrades performance in the 3- and 5-shot settings. This suggests that over-emphasizing textual deviation can suppress reliable visual reference matching when more normal references are available.

\subsection{Ablation on Backbones and Prompt Types}

\begin{table}[t]
    \centering
    \caption{Effect of fusion weights on anomaly detection performance.
    We fix $\alpha=1.0$ and vary $\beta$ in the dual-space anomaly score.
    Results are reported as average AUC-ROC (\%) and average category-wise standard deviation over 10 random runs under 1-, 3-, and 5-shot settings.}
    \label{tab:ablation_alpha_beta}
    \setlength{\tabcolsep}{5pt}
    \renewcommand{\arraystretch}{1.15}
    \small
    \begin{tabular}{ccccc}
        \toprule
        $\alpha$ & $\beta$
        & 1 sample
        & 3 samples
        & 5 samples \\
        \midrule
        1.0 & 0.00
        & 92.49 $\pm$ 5.64
        & 96.08 $\pm$ 1.97
        & 96.44 $\pm$ 1.43 \\
        
        1.0 & 0.25
        & 94.24 $\pm$ 4.09
        & 96.74 $\pm$ 1.48
        & 97.03 $\pm$ 1.02 \\

        1.0 & 0.50
        & 94.80 $\pm$ 3.41
        & \textbf{96.86} $\pm$ \textbf{1.22}
        & \textbf{97.12} $\pm$ \textbf{0.85} \\

        1.0 & 1.00
        & \textbf{94.89} $\pm$ 2.73
        & 96.41 $\pm$ 1.17
        & 96.63 $\pm$ 0.88 \\

        1.0 & 1.50
        & 94.54 $\pm$ \textbf{2.34}
        & 95.78 $\pm$ 1.28
        & 95.92 $\pm$ 1.05 \\
        \bottomrule
    \end{tabular}
\end{table}

\begin{table}[t]
    \centering
    \caption{Effect of CLIP backbones and prompt types on anomaly detection performance.
    We fix $\alpha=1.0$ and $\beta=0.5$ for all settings.
    Results are reported as average AUC-ROC (\%) and average category-wise standard deviation over 10 random runs under 1-, 3-, and 5-shot settings.}
    \label{tab:ablation_backbone_prompt}
    \setlength{\tabcolsep}{4pt}
    \renewcommand{\arraystretch}{1.15}
    \small
    \begin{tabular}{llccc}
        \toprule
        Backbone & Prompt Type
        & 1 sample
        & 3 samples
        & 5 samples \\
        \midrule
        ViT-B/16 & Name-only
        & 93.96 $\pm$ 3.68
        & 96.15 $\pm$ 1.54
        & 96.50 $\pm$ 1.30 \\

        ViT-B/16 & Template
        & 93.87 $\pm$ 3.66
        & 96.06 $\pm$ 1.46
        & 96.34 $\pm$ 1.27 \\

        ViT-B/16 & Semantic set
        & 93.51 $\pm$ 3.75
        & 95.74 $\pm$ 1.56
        & 96.04 $\pm$ 1.38 \\

        \midrule
        ViT-B/32 & Name-only
        & 94.44 $\pm$ 3.69
        & 96.70 $\pm$ 1.47
        & 96.94 $\pm$ 1.11 \\

        ViT-B/32 & Template
        & 94.78 $\pm$ 3.42
        & 96.85 $\pm$ \textbf{1.22}
        & 97.11 $\pm$ \textbf{0.85} \\

        ViT-B/32 & Semantic set
        & \textbf{94.80} $\pm$ \textbf{3.41}
        & \textbf{96.86} $\pm$ \textbf{1.22}
        & \textbf{97.12} $\pm$ \textbf{0.85} \\
        \bottomrule
    \end{tabular}
\end{table}
Table~\ref{tab:ablation_backbone_prompt} evaluates the effects of CLIP backbones and textual prompt designs. We compare ViT-B/16 and ViT-B/32, where ViT-B/32 consistently achieves slightly better results and is therefore adopted as the default backbone. For prompt design, we consider three variants: \textit{name-only}, which uses only the category name such as ``airplane''; \textit{template}, which uses a simple sentence template such as ``a photo of a \textit{class}''; and \textit{semantic set}, which combines multiple depth-aware and 3D-aware descriptions related to projected depth maps and point cloud objects. The results show that prompt design affects the quality of the textual normal space. Name-only prompts provide limited contextual information, while template prompts introduce a more natural language description. The semantic prompt set achieves the best overall performance with ViT-B/32, obtaining 94.80\%, 96.86\%, and 97.12\% average AUC-ROC under 1-, 3-, and 5-shot settings, respectively. This suggests that depth-aware and 3D-aware prompts better match the multi-view depth-map inputs and provide more effective semantic anchors. Therefore, we use ViT-B/32 with the semantic prompt set as the default configuration.

\subsection{Qualitative Analysis}
\begin{figure*}[tb]
    \centering
\includegraphics[width=1\linewidth]{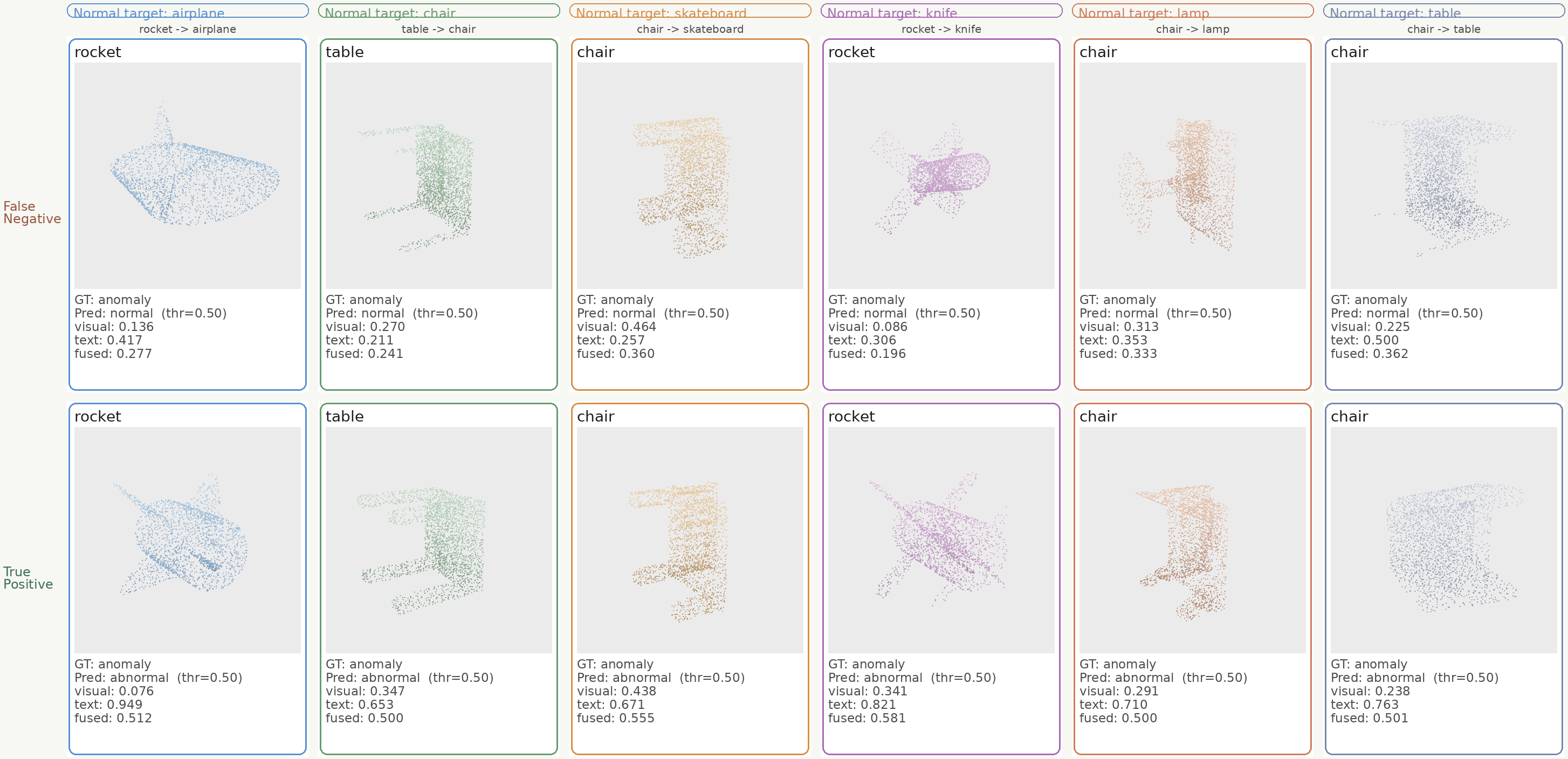}
\caption{Qualitative cross-category confusion examples under the 1-shot setting. Each column corresponds to one normal target category. The top row shows false negatives, where anomalous samples are incorrectly judged as normal, while the bottom row shows true positives, where anomalies are correctly detected.}
   \label{fig:2}
\end{figure*}

Fig.~\ref{fig:2} presents qualitative cross-category confusion examples under the 1-shot setting. Each column corresponds to one target normal category. The top row shows false negatives, where anomalous samples are incorrectly assigned low anomaly scores, while the bottom row shows true positives, where anomalies are correctly detected. For each sample, we visualize the visual score, text score, and fused score.

The qualitative results reveal two important observations. First, some false negatives remain challenging because both visual and textual cues are ambiguous. For instance, if an anomalous object has a global structure similar to the normal category, the visual branch may produce a low deviation score. At the same time, if the textual normal anchors are not sufficiently discriminative for this case, the semantic deviation may also be weak. Nevertheless, even in many false-negative cases, the text branch still contributes more abnormal evidence than the visual branch, although the combined score remains insufficient for correct detection. Second, in many true-positive cases, the text branch provides stronger abnormal evidence than the visual branch, leading to a higher fused anomaly score. This confirms that textual normal space alignment can complement visual reference matching and improve cross-category discrimination, especially when visual similarity alone is insufficient.

\section{Conclusion}
In this paper, we presented VT-3DAD, a training-free framework for few-shot cross-category 3D anomaly detection via visual-text normal space alignment. By combining visual deviation from few-shot references with semantic deviation from textual normal anchors, the proposed method improves both discrimination and robustness. Experiments on ShapeNetPart demonstrate that VT-3DAD achieves state-of-the-art performance under 1-, 3-, and 5-shot settings, with notable gains in the challenging 1-shot scenario. These results highlight the effectiveness of integrating visual and textual cues for training-free 3D anomaly detection. Future work will explore more adaptive prompt design and finer-grained geometric modeling.
\bibliographystyle{spiebib}
\bibliography{egbib}

\end{document}